\documentclass[conference]{IEEEtran}
\IEEEoverridecommandlockouts

\usepackage{cite}
\usepackage{amsmath,amssymb,amsfonts}
\usepackage{algorithmic}
\usepackage{graphicx}
\usepackage{textcomp}
\usepackage{xcolor}
\usepackage{url}
\usepackage{multirow}
\usepackage{booktabs}
\usepackage{listings}
\usepackage{xcolor}
\usepackage{fancyhdr}
\usepackage{array}
\usepackage{booktabs}
\usepackage{array}
\usepackage{longtable}
\usepackage{ragged2e} 
\usepackage[T1]{fontenc} 
\usepackage{adjustbox} 

\fancypagestyle{firstpage}{
  \fancyhf{}
  \fancyhead[C]{\small 2025 International Conference on Artificial Intelligence and Computing (AIC)}
}
\def\BibTeX{{\rm B\kern-.05em{\sc i\kern-.025em b}\kern-.08em
    T\kern-.1667em\lower.7ex\hbox{E}\kern-.125emX}}
\begin{document}
\title{\textbf{Advancing SLM Tool-Use Capability using Reinforcement Learning} \\
}

\author{
    \IEEEauthorblockN{Dhruvi Paprunia\textsuperscript{*}}
    \IEEEauthorblockA{
        Dept. of Computer Engineering\\
        MPSTME, NMIMS \\
        Mumbai-400056, India\\
        dhruvipaprunia1505@gmail.com
    }
    \and
    \IEEEauthorblockN{Vansh Kharidia\textsuperscript{*}}
    \IEEEauthorblockA{
        Dept. of Computer Engineering\\
        MPSTME, NMIMS \\
        Mumbai-400056, India\\
        vanshkharidia7@gmail.com
    }
    \and
    \IEEEauthorblockN{Dr. Pankti Doshi}
    \IEEEauthorblockA{
        Dept. of Computer Engineering\\
        MPSTME, NMIMS \\
        Mumbai-400056, India\\
        pankti.doshi@nmims.edu
    }
}

\maketitle
\thispagestyle{firstpage}

\begin{abstract}
In an era where tool-augmented AI agents are becoming increasingly vital, our findings highlight the ability of Group Relative Policy Optimization (GRPO) to empower SLMs, which are traditionally constrained in tool use. The ability to use tools effectively has become a defining feature of Large Language Models (LLMs), allowing them to access external data and internal resources. As AI agents grow more sophisticated, tool-use capabilities have become indispensable. While LLMs have made significant progress in this area, Small Language Models (SLMs) still face challenges in accurately integrating tool use, especially in resource-constrained settings.

This study investigates how Reinforcement Learning, specifically Group Relative Policy Optimization (GRPO), can enhance the tool-use accuracy of SLMs. By designing a well-defined reward system that reinforces structured JSON output, correct tool selection, and precise parameter usage, we demonstrate that GRPO enables SLMs to achieve significant improvements in tool-use capabilities (function calling/JSON output). Our approach provides a computationally efficient training method that enhances SLMs’ practical deployment in real-world AI applications.

\end{abstract}

\section{\textbf{Introduction}}
Large Language Models (LLMs) have progressed beyond simple text creation, and tool use has become increasingly important for complex, real-world tasks. Tool use in LLMs refers to their ability to utilize external resources such as APIs, databases, or software functions to extend their functionality beyond generating text. Tools are used for tasks such as performing calculations, making API calls to retrieve the current time and date, and more. This capability enables models to fetch real-time data, execute commands, or solve problems requiring dynamic interaction, making it indispensable for applications like AI agents in virtual assistants, robotic control, or automated workflows.
\\
However, while LLMs are usually  at tool use, their vast resource requirements and computation complexity restrict their use in every use-case. As a result, there is an increasing need for more compact and efficient Small Language Models (SLMs).
Small language models (SLMs) struggle in tool use compared to large language models (LLMs). As soon in Table 1. SLMs are typically trained on smaller, more specific datasets, resulting in a narrower knowledge base and limited contextual understanding compared to LLMs.
\\
This research addresses these challenges by using Reinforcement Learning (RL), specifically Group Relative Policy Optimization (GRPO), to enhance tool-use proficiency in SLMs. Unlike conventional fine-tuning approaches that require heavy computation and often lack adaptability, our method provides an efficient, effective solution that significantly boosts SLM tool-use accuracy, increasing their practical utility.

\section{\textbf{Related Works}}

\subsection{\textbf{Small Language Models (SLMs)}}
Small Language Models (SLMs) are a subset of language models that contain fewer parameters compared to Large Language Models (LLMs), which have hundreds of billions or trillions of parameters Lu et al. (2024)\textsuperscript{\cite{b2}}.

SLMs are designed to be more compact and efficient, typically ranging from 100M to 5B parameters, contrasting with LLMs that often require significant computational resources.

SLMs require less computational power and memory, making them ideal for deployment in resource-constrained environments such as edge devices and mobile applications. This efficiency is crucial for real-time processing Lu et al. (2024)\textsuperscript{\cite{b2}}. Furthermore, SLMs can be fine-tuned for specific domains, potentially leading to superior performance in targeted tasks. Wang et al. (2024)\textsuperscript{\cite{b3}} discuss how SLMs excel in specialized domains such as healthcare and law, where precision is paramount.

The reduced resource requirements lower training and inference costs, making AI development and deployment more accessible to smaller organizations\. Additionally, SLMs can be deployed in private cloud environments or on-premises, enhancing data protection. This is particularly critical for sectors handling sensitive information, as discussed in Wang et al. (2024)\textsuperscript{\cite{b3}}, where localized data handling is emphasized for privacy concerns.

Despite these advantages, SLMs face several challenges. CCompared to LLMs, SLMs have limited external knowledge and may struggle with more complex tasks \textsuperscript{\cite{b2}}. Furthermore, SLMs may struggle with context, leading to incorrect interpretations or outputs, a challenge highlighted by  Nguyen et al. (2023)\textsuperscript{\cite{b4}} in real-world applications. This can make them unsuitable for AI agents and tasks that require accurate contextual understanding. 

\subsection{\textbf{AI Agents}}
AI agents are autonomous or semi-autonomous software entities that can observe, plan, and act autonomously. These agents are often powered by language models, enabling advanced capabilities like reasoning and memory, as specified in Wang et al. (2024)\textsuperscript{\cite{b5}}. The concept dates back to the 1950s with early AI programs such as the Logic Theorist by Newell et al. (1959)\textsuperscript{\cite{b6}} and the development of Lisp in McCarthy et al. (1960)\textsuperscript{\cite{b7}}

The recent rise in AI agents is driven by advancements in language models, such as the GPT series introduced by in Radford et al. (2019)\textsuperscript{\cite{b8}}, enabling them to handle complex tasks   The integration of AI agents into daily life through applications like virtual assistants (Siri, Alexa) and autonomous vehicles, reflects their growing role across industries as shown in Shaik et al. (2019)\textsuperscript{\cite{b9}} 
 and Chen et al. (2024)\textsuperscript{\cite{b10}}.

As AI agents become integral to various domains, optimizing both accuracy and cost is crucial for practical deployment. Due to resource constraints, language models cannot be deployed in every scenario. Therefore, optimizing them for agentic workflows remains a critical challenge.

\subsection{\textbf{Tool Use in SLMs and Its Connection to AI Agents}}
Tool use refers to the ability of language models to interact with external tools or systems, such as APIs, to perform tasks beyond their inherent capabilities. This is a critical area of research, enhancing both AI agents and language models.

Tool use involves generating structured outputs, such as JSON, to invoke external functions\textsuperscript{\cite{b12}}. This extends capabilities to include accessing real-time data and controlling devices\textsuperscript{\cite{b13}}.
Tool use makes AI agents more versatile by providing access to real-time information and functionalities such as calculations. Qin et al. (2024)\textsuperscript{\cite{b13}} was one of the first approaches in finetuning LLMs in tool use.

AI agents use language models to understand and execute complex tasks efficiently. 
They can also use Small language models (SLMs) to plan and act, mirroring human reasoning processes to analyze tasks and formulate plans.
Additionally, AI agents can call on external tools or other agents to gather necessary information and execute specific tasks when needed, allowing them to tackle more complex goals autonomously.

\subsection{\textbf{Reinforcement Learning in LLMs, with a Focus on GRPO}}
Reinforcement learning (RL) fine-tunes LLMs by rewarding desired outputs, aligning them with human preferences  or specific tasks. 

RL enables LLMs to learn from feedback, improving their performance in tasks such as content generation. Ouyang et al. (2022)\textsuperscript{\cite{b14}} highlights its role in dynamic learning and goal alignment, enabling iterative improvements.

Group Relative Policy Optimization (GRPO), introduced in Shao et al. (2024)\textsuperscript{\cite{b15}}, and popularized by Deepseek-AI et al. (2025)\textsuperscript{\cite{b16}} is an efficient RL method, reducing memory and computational costs compared to Proximal Policy Optimization (PPO)\textsuperscript{\cite{b17}}. GRPO is an RL algorithm which replaces the critic neural network with a more efficient group-based advantage estimation. 

In the initial papers, GRPO was used to optimize LLM's math and reasoning capabilities. We adapt GRPO and its reward model to optimize tool-use in language models.

GRPO's efficiency makes it a promising direction for SLM fine-tuning, especially in resource-constrained settings. We use GRPO to fine-tune SLMs for tool use, showing significant accuracy improvements.

\subsection{\textbf{Research Gap Addressed}}
\begin{itemize}
    \item \textbf{Limited Reasoning Capacity}
        \paragraph{\textbf{Challenge}} SLMs often struggle to determine when a tool is needed or which tool to use without explicit guidance, due to their smaller size and reduced contextual understanding compared to LLMs.
        \paragraph{\textbf{Impact}} This can lead to missed opportunities for tool use or incorrect tool selection, reducing accuracy.

    \item \textbf{Hallucination and Over-Reliance}
        \paragraph{\textbf{Challenge}} SLMs may hallucinate tool outputs if a tool fails or isn’t available, or over-rely on tools even when their internal knowledge suffices.
        \paragraph{\textbf{Impact}} This reduces reliability and can confuse users with fabricated information.

    \item \textbf{Tool Dependency Risks}
        \paragraph{\textbf{Challenge}} SLMs can become overly dependent on tools, failing to leverage their internal knowledge even when it’s sufficient. This is amplified if training overly emphasizes tool use over standalone reasoning.
        \paragraph{\textbf{Impact}} Increased latency, unnecessary tool calls, and higher operational costs, particularly in environments where tool access is limited or expensive.

    \item \textbf{Ambiguity in User Intent}
        \paragraph{\textbf{Challenge}} SLMs often struggle to interpret ambiguous or underspecified user queries, making it hard to decide whether a tool is needed or which one to select without explicit cues.
        \paragraph{\textbf{Impact}} Misaligned tool usage (e.g., searching the web for a simple math problem) or failure to act, leading to user frustration.

    \item \textbf{Generalization Across Tools}
        \paragraph{\textbf{Challenge}} SLMs fine-tuned for specific tools (e.g., a weather API) may struggle to adapt to new or similar tools (e.g., a different weather API with a slightly different format) without retraining.
        \paragraph{\textbf{Impact}} Reduced flexibility and higher maintenance costs when scaling to new use cases or replacing deprecated tools.
\end{itemize}

\subsection{\textbf{Solution}}
To address these challenges, we developed a novel reward model, optimized with Group Relative Policy Optimization (GRPO), for fine-tuning Small Language Models. This approach trains SLMs to master structured tool use—ensuring valid JSON output, precise tool selection, and accurate parameter specification. It also penalizes extra text with a strict zero-reward mechanism. This efficient, adaptable solution significantly elevates SLM tool-use accuracy, making them reliable and scalable for real-world applications.

\section{\textbf{Methodology}}

\subsection{\textbf{Experimental Setup}}
The experiments conducted for this research, as detailed in the Reward Modelling Section. The experiments were conducted on a T4 GPU. The proposed model was fine-tuned and evaluated using the SalesforceXLAM\_60K\textsuperscript{\cite{b1}} dataset. The primary goal is to enhance tool-use accuracy in SLMs by optimizing tool-use capabilities through reinforcement learning.

\subsection{\textbf{Dataset Description}}
We used the SalesforceXLAM dataset to evaluate the tool-use capabilities of our models. Figure 1 shows a sample from the dataset and highlights the key columns:

\begin{figure}
    \centering
    \includegraphics[width=1\linewidth]{"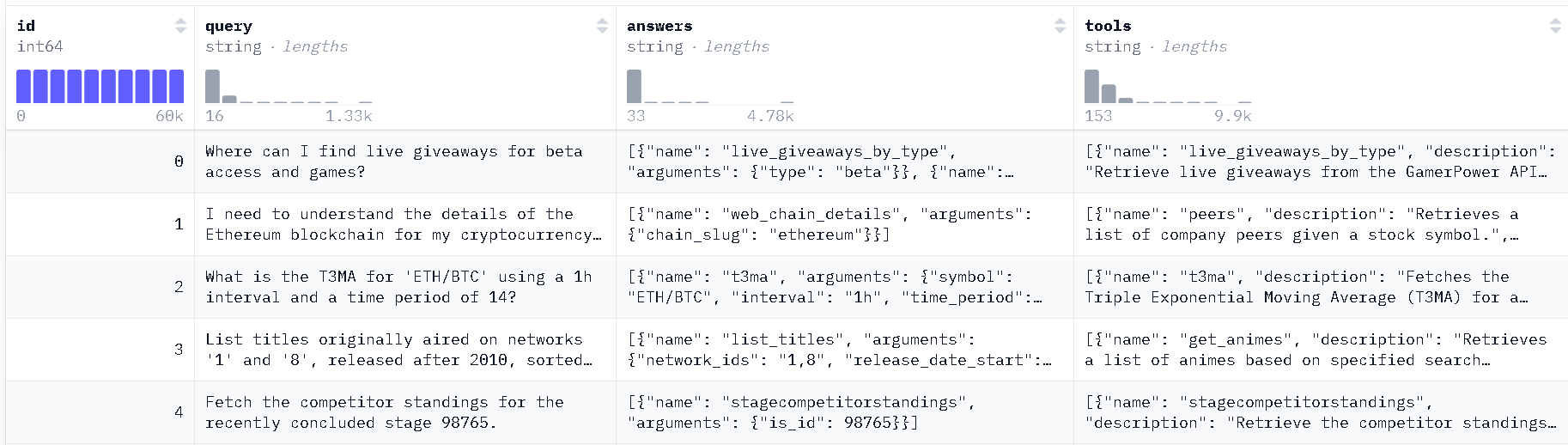"}
    \caption{Dataset Snapshot}
\end{figure}

\begin{itemize}
    \item \textbf{id}: Unique identifier for each query.
    \item \textbf{query}: User input requiring function execution.
    \item \textbf{answers}: Expected model-generated response in JSON format.
    \item \textbf{tools}: A list of available tools from which the LLM must select the appropriate one.
\end{itemize}
A subset of this dataset was used for training and evaluation:

\begin{itemize}
    \item \textbf{Training set}: 4,000 samples, ensuring diverse function-call scenarios.
    \item \textbf{Testing set}: 1,000 samples, used to evaluate model performance.
\end{itemize}

\section{\textbf{Algorithm}}

The primary challenges in this setting involve ensuring the model learns to: 
(1) produce valid JSON, 
(2) select the correct tool names, and 
(3) provide accurate arguments. 
Additionally, a significant issue observed in model outputs was the presence of extraneous text outside the structured format, which necessitated a strict penalty mechanism. As shown in Table 1.

\subsection{\textbf{Progressive Learning of Model Capabilities}}
Rather than designing a complex reward function from the outset, we employed an incremental approach to reward modeling, wherein we analyzed the model’s learning trajectory at different stages. Initially, the model was trained to output well-formed JSON, which it learned relatively quickly. Because JSON formatting presented less of a challenge, we reduced the reward for valid JSON output (from $0.5$ to $0.125$) to prioritize learning tool selection and argument correctness.

Next, we separately trained the models such as Qwen 2.5 1.5B, Qwen 2.5 3B, Llama 3.2 3B to learn tool names and tool arguments. Through this process, we observed that both were comparably difficult for the model to learn. The model exhibited a tendency to make minor hallucination errors, especially in argument names. Function names, on the other hand, proved to be more stable. Consequently, we assigned a slightly higher reward to argument correctness than function name correctness to reflect its relative difficulty.

A persistent challenge across all language models was the inclusion of extraneous text, where models would generate correct tool uses but include additional text, such as:

\begin{quote}
\textit{"This is the correct tool call: \{ ...valid JSON... \}"}
\end{quote}

To counter this, we implemented a strict penalty mechanism: any extraneous text, whether preceding or following the JSON output, resulted in a reward of zero. This strict enforcement ensured that the model learned to produce only the JSON output and nothing else.

We refer to this iterative refinement of the reward function as \textbf{capability-aware reward modeling}, because the reward structure evolves based on the model's observed learning behavior.
\subsection{\textbf{Formal Definition of the Reward Function}}

The formulas below were created to measure how accurately the model generates tool calls.

\subsubsection{\textbf{Extraneous Text Penalty}}
To enforce strict adherence to the expected structured output, the reward function applies an immediate penalty if the predicted answer contains any content beyond a valid JSON structure. If extraneous text is present, the reward is set to zero:
\begin{equation}
R_{\text{final}} = 0, \quad \text{if extraneous text exists}
\end{equation}
Similarly, a prediction that is empty or contains invalid JSON also receives a reward of zero.

\subsubsection{\textbf{JSON Validity Reward}}
Once the model learned to produce valid JSON reliably, we reduced the JSON validity reward to a minor component, set at:
\begin{equation}
R_{\text{json}} = 0.125
\end{equation}
If the output is invalid JSON, this component is set to zero.

\subsubsection{\textbf{Function Name Reward}}
Correct tool selection is evaluated based on whether the predicted function names match the expected tool calls. The function name reward is distributed among the expected tool use as follows:
\begin{equation}
R_{\text{fn}} = \frac{0.375}{\max(n_{\text{expected}}, 1)} \times n_{\text{correct functions}}
\end{equation}
where $n_{\text{expected}}$ is the number of expected tool calls, and $n_{\text{correct functions}}$ is the number of correctly predicted tool names.

\subsubsection{\textbf{Argument Matching}}
As tool argument correctness was slightly more difficult for the model to learn than function names, we allocated a higher reward weight (0.5) to argument correctness. The argument reward is computed as:
\begin{equation}
R_{\text{args}} = \frac{0.5}{\max(n_{\text{expected}}, 1)} \times \sum_{i=1}^{n_{\text{correct functions}}} A_i
\end{equation}
where $A_i$ represents the argument correctness score for each correctly predicted tool use:
\begin{equation}
A_i = \frac{\text{number of correctly predicted arguments}}{\text{total number of expected arguments}}
\end{equation}

\subsubsection{\textbf{Penalty for Extra Tool Calls}}
To prevent models from over-generating tool calls, a scaling penalty was introduced. If the model predicts more tool calls than expected, the function name and argument rewards are scaled down using:
\begin{equation}
\text{Scaling Factor} = \frac{1}{\frac{n_{\text{predicted}}}{\max(n_{\text{expected}}, 1)}}
\end{equation}
The adjusted rewards are then computed as:
\begin{equation}
R_{\text{fn}}' = R_{\text{fn}} \times \text{Scaling Factor}
\end{equation}
\begin{equation}
R_{\text{args}}' = R_{\text{args}} \times \text{Scaling Factor}
\end{equation}
where $n_{\text{predicted}}$ is the number of tool calls generated by the model. This penalty discourages over-generation of tool calls while ensuring accurate predictions remain fairly rewarded.

\subsubsection{\textbf{Final Reward Calculation}}
The final reward is computed as:
\begin{equation}
R_{\text{final}} = \begin{cases} 0, & \text{if extraneous text exists} \\ R_{\text{fn}}' + R_{\text{args}}' + R_{\text{json}}, & \text{otherwise} \end{cases}
\end{equation}

This formulation balances correctness, structure, and conciseness while strictly penalizing errors that impact tool usability.
\begin{figure}
    \centering
    \includegraphics[width=0.75\linewidth]{"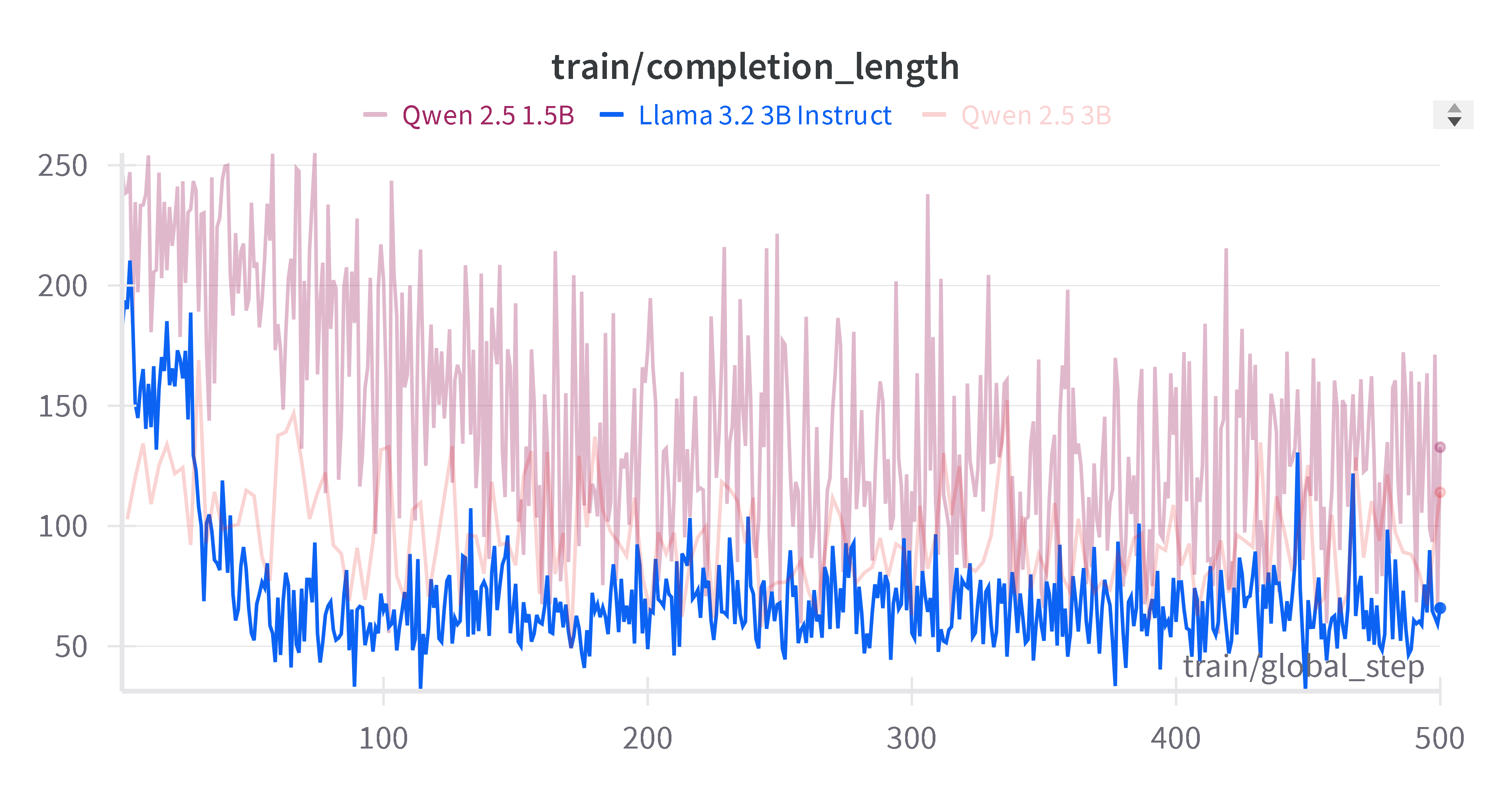"}
    \caption{Completion Length}
    \label{fig:enter-label}
\end{figure}

\section{\textbf{Reward Modelling}}

\subsection{\textbf{Why Complex Reward Models Failed}}
Initial experiments using more complex reward models that incorporated soft penalties for minor errors, rather than strict zero-reward penalties, led to several undesirable behaviors:
\begin{figure}
    \centering
    \includegraphics[width=0.75\linewidth]{"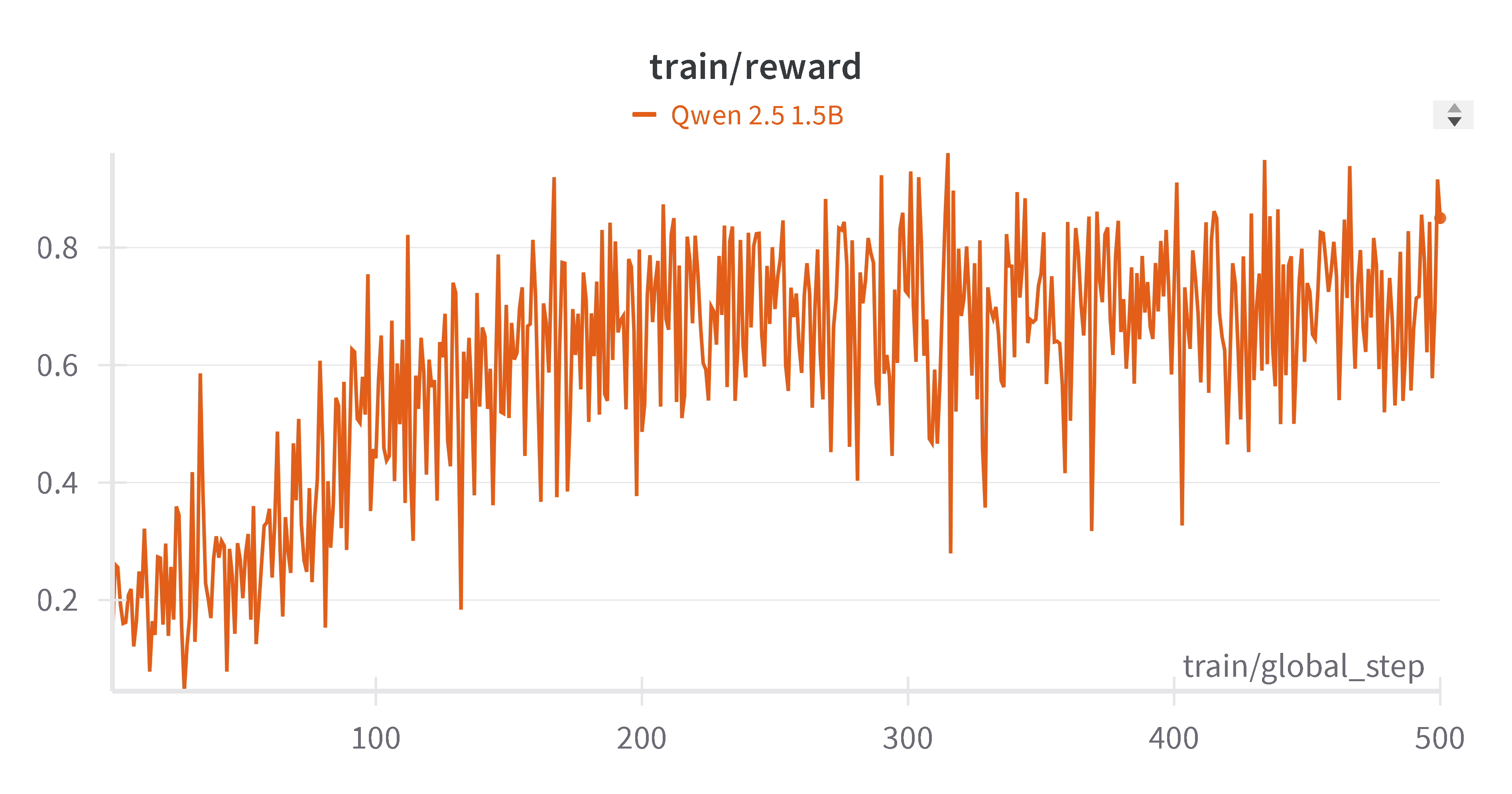"}
    \caption{Qwen 2.5 1.5B Reward}
\end{figure}
\begin{itemize}
    \item \textbf{Persistent Extraneous Text Generation:} When extraneous text received a soft penalty instead of a zero reward, the model continued to produce additional explanations or formatting, even when it had correctly generated the JSON output.
    \item \textbf{Instability in Argument Predictions:} Soft penalties on incorrect arguments resulted in the model assigning arbitrary values when uncertain, leading to unpredictable errors rather than complete failures that could be easily debugged.
    \item \textbf{Over-generation of Tool Calls:} Without a strict penalty, models tended to predict additional tool uses to hedge their outputs, as redundant calls were still slightly rewarded.
\end{itemize}

These findings reinforced the necessity of a strict, structured reward function rather than a softer reward shaping approach.

\begin{figure}
    \centering
    \includegraphics[width=0.75\linewidth]{"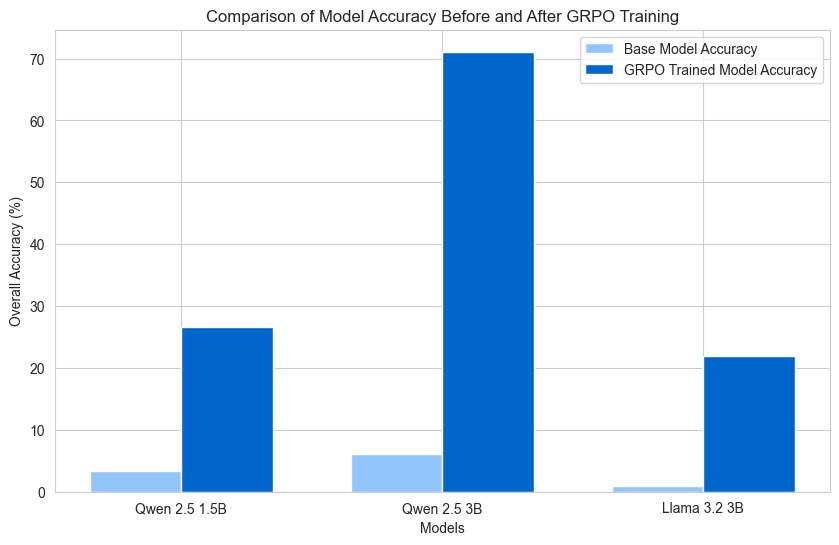"}
    \caption{Comparison of Models Before and After GRPO training}
\end{figure}
\subsection{\textbf{Why Capability-Aware Reward Modelling}}
Through capability-aware reward modeling, we progressively refined the reward function based on the model's observed learning behavior. This approach ensured that the model first mastered JSON formatting before focusing on tool selection and argument correctness. By strictly penalizing extraneous text and over-generation while assigning slightly higher importance to argument correctness, we achieved a stable and effective reward model for tool use in language models.

\subsection{\textbf{Our ‘Aha’ Moment}}
Initially, the model tended to output extraneous text along with the JSON output. However, the aggressive 'zero reward if extraneous text' policy quickly taught the model to avoid extraneous text. As shown in the graph below in Fig 2, immediately impacting one of the key reasons for failure in LLM tool use.

\section{\textbf{Evaluation}}
\subsection{\textbf{Model Training and Fine-tuning}}

\subsubsection{\textbf{Fine-tuning Configuration}}
\textbf{LoRA-based Parameter-Efficient Fine-tuning:}
\begin{itemize}
    \item Rank: 64
    \item Low-Rank Adaptation (LoRA) allows fine-tuning with minimal computational overhead while retaining most of the performance.
\end{itemize}
To optimize function-calling accuracy, we fine-tuned multiple small language models (LLMs), including Qwen 2.3 3B, Qwen 2 1.5B Instruct, and Llama 3.2 3B Instruct, using Group Relative Policy Optimization (GRPO). The objective was to improve structured tool invocation, response formatting, argument correctness, and implicit penalty enforcement to avoid extra functions in the output.

\begin{table}[htbp]
\centering
\label{tab:model_comparison}
\begin{tabular}{|p{0.27\columnwidth}|p{0.32\columnwidth}|p{0.27\columnwidth}|}
\hline
\textbf{Expected Output} & \textbf{Base Model Output} & \textbf{Finetuned Output (Qwen 2.5 1.5B)} \\
\hline
[\{name: qr\_code\_image, arguments: \{size: 7, url: example.com\}\}, \{name: ec, arguments: \{password: Secure123, penalty: 0.3, format: json\}\}]
&
[\{name: qr\_code\_image\_generator, arguments: \{url: example.com\}\}, \{name: ec, arguments: \{password: Secure123, penalty: 0.3, format: json\}\}]
&
[\{name: qr\_code\_image, arguments: \{size: 7, url: example.com\}\}, \{name: ec, arguments: \{password: Secure123, penalty: 0.3, format: json\}\}]
\\
\hline
\end{tabular}
\vspace{2mm}
\caption{Comparison of Model Outputs}
\end{table}

\subsubsection{\textbf{Training Hyperparameters}}
\begin{itemize}
    \item Learning rate: 5e-6 (optimized for stability and convergence)
    \item Learning rate scheduler: cosine\_with\_restarts (ensures the learning rate does not diminish permanently, beneficial for RL-based fine-tuning)
    \item Optimizer: AdamW (8-bit) for efficient memory utilization
    \item Gradient accumulation steps: 4 (improves stability and enables effective batch processing)
    \item Number of generations per training query: 8 (GRPO benefits from a higher number of sampled generations, minimum 8 generations needed for GRPO to be effective)
    \item Maximum completion length: 8192 tokens (ensures the model can handle long tool use sequences, so that the output doesn’t end abruptly)
    \item Total training steps: 500 (due to compute limitations)
    \item vLLM-based inference acceleration enabled to enhance training speed while maintaining accuracy.
\end{itemize}

\subsection{\textbf{Evaluation Metrics}}
To measure the model’s performance, we used multiple evaluation metrics focusing on structured tool uses:

\begin{itemize}
    \item \textbf{Overall Accuracy:} Measures the correctness of the predicted function name, arguments, and JSON structure.
    \item \textbf{JSON Validity:} Evaluates whether the response is syntactically valid JSON, ensuring structural consistency.
\end{itemize}

\subsection{\textbf{Sample Output}}

Table 1 shows a sample tool call by the base (untrained) model and a finetuned (trained) model.

The base model output uses the wrong tool name - \texttt{qr\_code\_image\_generator} instead of \texttt{qr\_code\_image}, and is missing the \texttt{size} argument. The finetuned model, on the other hand, generates the perfect tool call.

\subsection{\textbf{Evaluated Models}}

We use instruction-tuned Qwen 2.5 1.5B, Qwen 2.5 3B\textsuperscript{\cite{b20}}, and Llama 3.2 3B\textsuperscript{\cite{b21}}, which is also depicted in Fig 3.\\

\begin{table}[h]
    \centering
    \begin{tabular}{|l|c|c|}
        \hline
        \textbf{Qwen 2.5 1.5B} & \textbf{JSON Validity} & \textbf{Overall Accuracy} \\
        \hline
        Base Model & 73\% & 3.40\% \\
        GRPO Trained Model & 100\% & 26.60\% \\
        \hline
    \end{tabular}
    \label{tab:qwen_15b}
    \vspace{2mm}
    \caption{Qwen 2.5 1.5B Performance}
\end{table}
Table 2 showed an approximately 6.28$\times$ increase in accuracy after using the GRPO Trained Model. An example of the reward function for Qwen 2.5 1.5B Model is show in Fig 3.\\
\begin{table}[h]
    \centering
    \begin{tabular}{|l|c|c|}
        \hline
        \textbf{Qwen 2.5 3B} & \textbf{JSON Validity} & \textbf{Overall Accuracy} \\
        \hline
        Base Model & 100\% & 6.10\% \\
        GRPO Trained Model & 100\% & 71.1\% \\
        \hline
    \end{tabular}
    \label{tab:qwen_3b}
    \vspace{2mm}
    \caption{Qwen 2.5 3B Performance}
\end{table}

Table 3 showed an approximately 10.65$\times$ increase in accuracy after using the GRPO Trained Model.\\
\begin{table}[h]
    \centering
    \begin{tabular}{|l|c|c|}
        \hline
        \textbf{Llama 3.2 3B} & \textbf{JSON Validity} & \textbf{Overall Accuracy} \\
        \hline
        Base Model & 73.2\% & 0.98\% \\
        GRPO Trained Model & 100\% & 22.00\% \\
        \hline
    \end{tabular}
    \label{tab:llama_3b}
    \vspace{2mm}
    \caption{Llama 3.2 3B Performance}
\end{table}

Table 4 showed an approximately 21.65$\times$ increase in accuracy after using the Llama GRPO Trained Model.

\section{\textbf{Conclusion}}
This research conclusively demonstrates Group Relative Policy Optimization (GRPO) as a highly effective and compute-efficient method for boosting tool-use in Small Language Models (SLMs). GRPO fine-tuning led to dramatic accuracy improvements (6x-21x) and yielded near-perfect JSON output across the tested SLMs. Our capability-aware reward modeling, particularly the strict penalty for extraneous text, was key to success. GRPO effectively empowers resource-constrained SLMs to achieve significant tool-use proficiency, narrowing the gap with larger models and expanding their real-world applicability.

\subsection{\textbf{Real World Impact}}
This research has significant real-world implications by directly addressing the limitations of Small Language Models (SLMs) in tool use, a crucial capability for practical AI applications. By enabling SLMs to effectively leverage external tools and APIs through GRPO fine-tuning, this work paves the way for more efficient and deployable AI agents in resource-constrained environments. This is particularly relevant for scenarios where large language models are impractical due to computational cost or latency, such as edge computing devices, mobile applications, and embedded systems. Improved tool-use in SLMs can unlock a wider range of functionalities in these settings, from intelligent personal assistants on smartphones to automated systems in resource-limited industries, making sophisticated AI more accessible and broadly applicable.

Furthermore, the enhanced accuracy and reliability of tool-use achieved through GRPO have the potential to significantly improve the user experience and trustworthiness of AI-driven applications. By ensuring precise tool use and structured JSON output, the method reduces errors and hallucinations, leading to more predictable and dependable AI agents. This increased reliability is crucial for real-world adoption, especially in domains where accuracy is paramount, such as data analysis, automation of business processes, and integration with critical infrastructure. Ultimately, by democratizing access to effective tool-augmented AI through efficient SLMs, this research contributes to a future where intelligent and helpful AI assistants are more readily available and integrated into everyday life, even in resource-limited contexts.

\subsection{\textbf{Future Work}}
Future research should focus on several key areas. Exploring the scalability of GRPO to larger language models and extended training durations is important. Further refining the reward functions to encompass more nuanced aspects of tool selection and automating reward design would be beneficial. A comparative analysis of GRPO against other Reinforcement Learning methods specifically for tool-use is warranted. Evaluating GRPO's effectiveness with more diverse and complex real-world tools and APIs should also be pursued.

\end{document}